\documentclass{article}
\usepackage{evoworld_arxiv,times}

\usepackage{amsmath,amsfonts,bm}

\def\eqref#1{equation~\ref{#1}}

\def\1{\bm{1}}

\DeclareMathAlphabet{\mathsfit}{\encodingdefault}{\sfdefault}{m}{sl}
\SetMathAlphabet{\mathsfit}{bold}{\encodingdefault}{\sfdefault}{bx}{n}

\usepackage{hyperref}
\usepackage{url}
\usepackage{graphicx}
\usepackage{booktabs}
\usepackage{multirow}
\usepackage[table]{xcolor}
\usepackage{tikz}
\usepackage[capitalize]{cleveref}
\usepackage{listings}
\usepackage{caption}
\usepackage{subcaption}
\usepackage{wrapfig}
\usepackage{algorithm}
\usepackage{xspace}
\usepackage{dsfont}
\usepackage{pifont}
\usepackage{tcolorbox}
\definecolor{frontmatterbg}{HTML}{E8F2FC}
\definecolor{frontmattertext}{HTML}{20334F}
\definecolor{projectlink}{HTML}{1877D2}

\definecolor{todocolor}{RGB}{184,134,11}
\definecolor{oursrow}{RGB}{232,243,255}

\newcommand{\ours}{EvoWorld\xspace}

\newcommand{\evoworldmark}{%
  \tikz[x=0.034ex,y=0.034ex,baseline=-0.55ex]{%
    \draw[line width=0.136ex,draw={rgb,255:red,241;green,169;blue,78}]
      (-23,0) .. controls (-15,11) and (-7,16) .. (0,16)
      .. controls (7,16) and (15,11) .. (23,0)
      .. controls (15,-11) and (7,-16) .. (0,-16)
      .. controls (-7,-16) and (-15,-11) .. cycle;
    \fill[fill={rgb,255:red,143;green,215;blue,232}] (0,0) circle (9);
    \draw[line width=0.068ex,draw={rgb,255:red,7;green,19;blue,29}]
      (0,9) .. controls (-3.2,3.4) and (-3.2,-3.4) .. (0,-9)
      (0,9) .. controls (3.2,3.4) and (3.2,-3.4) .. (0,-9)
      (-8,0)--(8,0);
  }%
}
\title{EvoWorld: Towards Out-of-Sight World Evolution in Generative Video Models}

\author{
\mdseries
\begin{tabular}{c}
Yequan Bie\textsuperscript{1},
Dengyang Jiang\textsuperscript{1},
Zhixuan Chen\textsuperscript{1},
Xiaozhuan Liang\textsuperscript{2}, \\
Haoxuan Che\textsuperscript{1},
Quande Liu\textsuperscript{3,\ding{41}},
Xi Chen\textsuperscript{2},
Hao Chen\textsuperscript{1,\ding{41}} \\
\textsuperscript{1}The Hong Kong University of Science and Technology \\
\textsuperscript{2}Tencent \quad
\textsuperscript{3}The Chinese University of Hong Kong \\
\href{https://tommy-bie.github.io/projects/evoworld/}{\textcolor{projectlink}{%
\evoworldmark\enspace\texttt{https://tommy-bie.github.io/projects/evoworld/}}}
\end{tabular}
}

\iclrfinalcopy
\makeatletter
\def\@maketitle{\vbox{\hsize\linewidth
  {\centering\LARGE\normalfont\bfseries\@title\par}
  \vskip 18pt
  \hbox to\linewidth{\hfil{\normalfont\@author}\hfil}
  \vskip 16pt}}
\makeatother
\begin{document}
\thispagestyle{plain}

\begin{tcolorbox}[
  colback=frontmatterbg,coltext=frontmattertext,
  boxrule=0pt,arc=12pt,boxsep=0pt,
  left=14pt,right=14pt,top=18pt,bottom=18pt,
  before skip=0pt,after skip=18pt]
\maketitle
\renewenvironment{abstract}{%
  \par\begin{center}\large\bfseries Abstract\end{center}
  \noindent\ignorespaces
}{\par}

\begin{abstract}
Generative video models are increasingly used as visual world simulators, yet their primary strength lies in rendering plausible pixels rather than reasoning about how a world should evolve. Coherent world evolution requires more than visual fidelity: when an event occurs, a model must infer its consequences for the entities in the world and then render the resulting state. For example, after a wild beast appears, the model should not only depict the wild beast, but also infer whether a nearby person should flee, take cover, or fight back based on the realized scene. Existing diffusion-based video models lack an explicit mechanism for maintaining world state and reasoning over such event-conditioned consequences, often producing visually plausible but behaviorally static or causally inconsistent outputs. To make this reasoning gap concrete and measurable, we propose a new problem, \emph{Out-of-Sight Causal Evolution}, which requires a model to render an event, reason about its consequences for the evolving world state, and reveal a causally plausible outcome as the world continues to evolve, even out of sight. As a baseline reference solution, we introduce EvoWorld, a brain--renderer collaboration framework in which a vision-language model serves as the world brain: it observes generated segments, maintains an evolving account of the world, and reasons about plausible state transitions, while a segment-autoregressive video renderer realizes them visually. We further introduce EvoWorld-Data and EvoWorld-Bench for training and evaluating event realization, causal reactions, and revisit consistency. Experiments show that EvoWorld substantially improves event-conditioned causal evolution over recent models while preserving visual quality and consistency, establishing a reference point for reasoning-guided, autonomous world evolution beyond the current field of view.

\end{abstract}
\end{tcolorbox}
\begingroup
\renewcommand{\thefootnote}{\mbox{\ding{41}}}
\footnotetext[1]{Corresponding authors.}
\endgroup

\section{Introduction}\label{sec:intro}

Building models that can simulate evolving environments has become an important research direction in generative AI. By learning world dynamics, a world model uses observed history and input actions to predict plausible future states \citep{lingbot-world-v1,lingbot-world-v2,zhang2026echowm,vjepa2}. Such predictive representations have enabled progress in decision making and autonomous driving \citep{survey_wm_autonomous_driving, zhou2025hermes,liang2025unifuture,gao2024vista,gao2024magicdrive}, agent training and embodied intelligence \citep{survey_wm_embody_ai,wang2025embodiedgen,hafner2025training_agent}, and synthetic environment generation \citep{bruce2024genie,sora}.

Generative video models \citep{wan2025wan,hacohen2026ltx-2,yang2025cogvideox,wu2025hunyuanvideo1.5} provide a direct approach to visual world modeling \citep{lingbot-world-v1,he2025matrix2,team2026alayaworld}: their predictions are immediately observable, while text and camera controls offer intuitive interaction. Yet locally plausible frames alone do not yield a coherent long-horizon world. A model must preserve the environments it has traversed and the events that have occurred; otherwise, later states become disconnected from the history that should have produced them.
Accordingly, recent video world models have explored various mechanisms for long-horizon memory and consistency, including recurrent state-space memory \citep{zhu2026sana-wm}, memory retrieval or dynamic frame eviction \citep{yu2025contex_as_memory,yang2026stableworld}, and geometry-grounded spatial memory (e.g., 3D point clouds) \citep{wu2026video_pointcloud,team2026alayaworld,ma2026closing}.

Most existing approaches nevertheless focus on preserving or retrieving previously observed visual states. Such persistence is necessary but insufficient: coherent world evolution also requires reasoning about how entities should change as new events unfold. This highlights a key limitation of current video models: they primarily rely on diffusion-based generation to render visual observations, without an explicit mechanism for reasoning over world states and event consequences. For example, when a monster appears, a model should not only render the monster faithfully, but also infer whether a nearby character should flee, seek cover, or fight back according to the realized world state. The challenge becomes especially pronounced when the affected subject is outside the current field of view: its resulting state cannot be determined simply by restoring its last observed appearance, but must also incorporate changes inferred from the intervening event. To make this reasoning gap concrete and measurable, we formulate a new problem, \emph{Out-of-Sight Causal Evolution}, which requires a model to realize an event, reason about how the event changes the relevant world state, including its consequences for an out-of-sight subject, and render the resulting state when the subject is revisited.

\begin{figure*}[t]
\centering
\includegraphics[width=\textwidth]{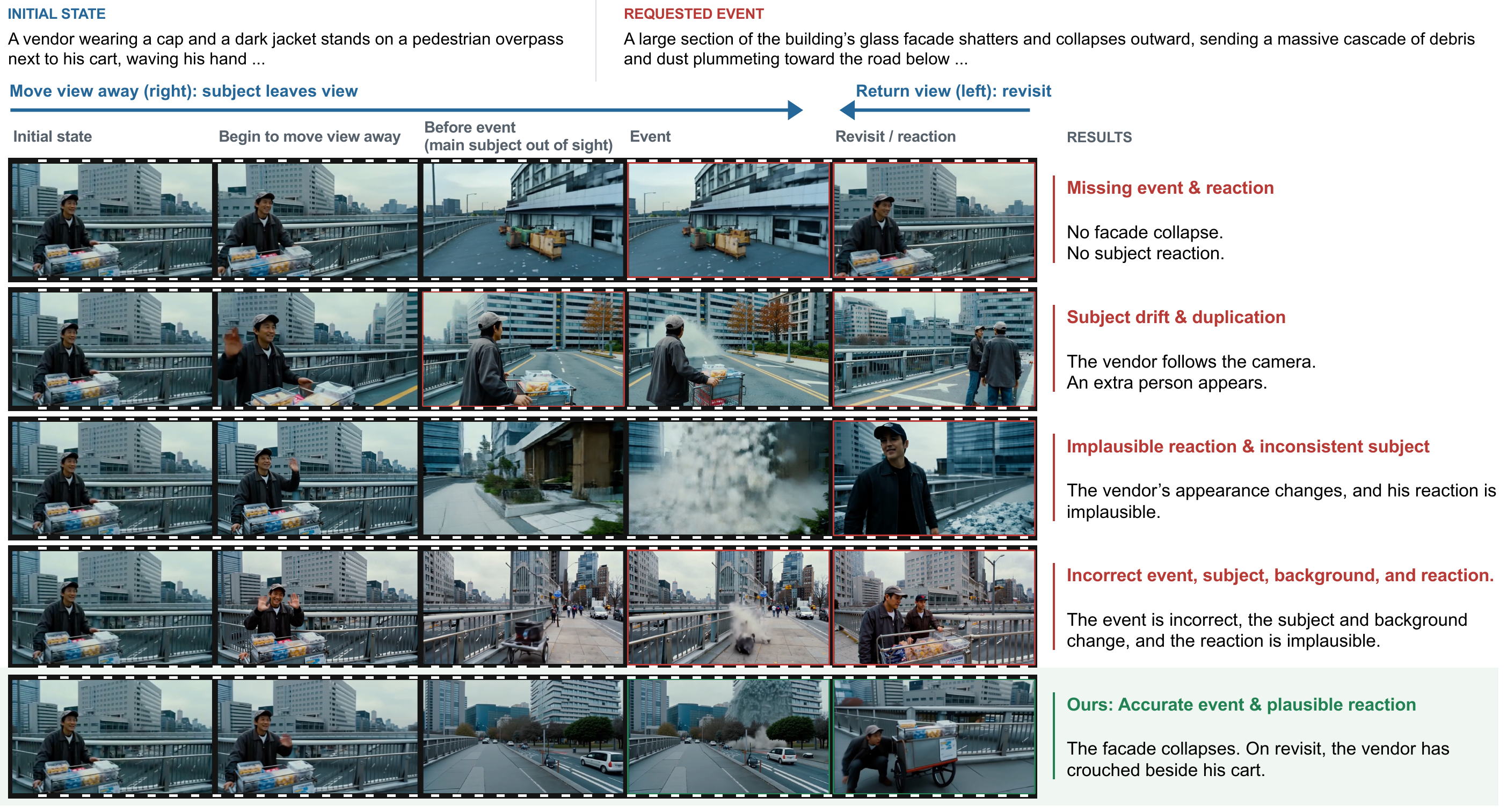}
\vspace{-0.6cm}
\caption{Challenges in \textit{Out-of-Sight Causal Evolution}. The camera first observes a subject (a vendor and his handcart), moves away to reveal a newly requested event (a collapsing building facade), and then moves back to the original scene. Previous methods (rows 1--4) exhibit various failures. In contrast, \ours (row 5) realizes the event, revisits the original subject, and renders a causally plausible reaction while preserving subject and scene consistency.}
\label{fig:challenge_compare}
\vspace{-0.7cm}
\end{figure*}

\emph{Out-of-Sight Causal Evolution} couples three challenges that existing models rarely address simultaneously: preserving subject and scene consistency across camera motion; realizing an event introduced by a mid-rollout prompt switch; and reasoning about, then rendering, a causally plausible world-state transition, revealed through the subject's response when the camera returns. As illustrated in \cref{fig:challenge_compare}, HY-World 1.5 \citep{hyworld1.5}, Echo-WM \citep{zhang2026echowm}, LingBot-World \citep{lingbot-world-v1}, and SANA-WM \citep{zhu2026sana-wm} exhibit various failures: the event or reaction may be absent, the subject may follow the camera or become duplicated, appearance and background may drift, or the subject may exhibit implausible behavior. Appearance persistence, instruction following, and event-conditioned behavior must therefore be measured jointly. Therefore, we introduce EvoWorld-Data, a controlled generative dataset of around 12k long videos ($\geq 10$ seconds). The training set spans 19 event families and 65 fine-grained event types, 16 scene families and 63 fine-grained scene types, and 14 reaction types across photorealistic and game-world domains. Segment-level prompts and temporal annotations provide direct supervision for event realization, long-range visual continuity, and behavioral change.

As a reference solution, we develop EvoWorld, a reasoning-and-retrieval-augmented framework that separates world-state reasoning from visual rendering. A VLM serves as a \textit{world brain} that continuously perceives and records the evolving world state, while the \textit{renderer} generates the next segment based on the reasoning from the \textit{world brain}.
We train the renderer on EvoWorld-Data with Identity-Anchored Attention and a two-stage strategy that first learns reactive rendering from clean data and then reduces exposure bias through online autoregressive rollout training.
We further introduce EvoWorld-Bench, which evaluates whether the camera revisits the original region, the switched prompt realizes the requested event, the subject responds plausibly, and the subject and background remain consistent. These coupled capabilities are not captured by existing video benchmarks such as VBench \citep{huang2024vbench}, which primarily emphasize perceptual quality and short-range temporal behavior.

Our contributions are threefold: (i) We formulate \emph{Out-of-Sight Causal Evolution}, an event-conditioned world-evolution problem that requires revisit consistency, mid-rollout event realization, and causal reactive evolution. (ii) We introduce EvoWorld, a world-brain--renderer collaboration framework that separates world-state reasoning from visual rendering to continuously perceive, reason about, and render out-of-sight world dynamics. (iii) We propose EvoWorld-Data, which contains around 12k videos spanning diverse scenes, events, and reactions, and introduce EvoWorld-Bench to measure the \emph{Out-of-Sight Causal Evolution} capability. We hope this work holds the potential to advance the research of out-of-sight dynamics and causal evolution in video world modeling.

\section{Related Work}\label{sec:related}
\vspace{-0.2cm}
\textbf{Video world models \& autoregressive video generation.}
Generative video backbones \citep{wan2025wan,hacohen2026ltx-2,minimax2026h3} provide the visual synthesis foundation for interactive world simulation, including recent models and systems such as XGEN \citep{xgen}, Zing-0.5 \citep{zing0.5}, and WorldPlay2 \citep{zhang2026worldplay2}. LingBot-World \citep{lingbot-world-v1} and LingBot-World-v2 \citep{lingbot-world-v2} support long-horizon, real-time interaction, with the latter combining bidirectional and autoregressive attention. Echo-WM \citep{zhang2026echowm} targets open and enterable omnimodal world simulation, while Alaya-EVOKE \citep{yin2026alaya-evoke} combines persistent world-state memory, mid-rollout reprompting, and few-step generation. Segment-autoregressive methods preserve full attention within each generated segment to reduce the quality degradation associated with fully causal adaptation. YUME-1.5 \citep{mao2026yume1.5} combines long-context generation with bidirectional-attention distillation. Matrix-Game~3.0 \citep{wang2026matrixgame3} uses memory-augmented multi-segment autoregressive distillation. BiWM \citep{rui2026biwm} provides a general recipe for adapting bidirectional video backbones into controllable segment-autoregressive world models. We build on this formulation and focus on whether the generated world continues to evolve beyond the current field of view.

\textbf{Out-of-sight memory and dynamics.}
Long-horizon consistency does not ensure that an unobserved world continues to evolve. Recent evaluations expose this limitation through camera look-away, occlusion, or illumination changes \citep{ma2026out_stevo_bench,lu2026current_lack_persistent_state_core_survey}. ReMind \citep{xu2026teaching_remind} trains a causal video generator to recover evolving states across interruptions. LiveWorld \citep{duan2026liveworld} maintains a persistent global state by separating a static 3D environment from dynamic entities advanced by an evolution engine. Hybrid Memory \citep{chen2026out_hm_world} distinguishes static-background preservation from dynamic-subject tracking across exit--reentry intervals. These works primarily examine whether an active process or motion continues through an observation gap. Our setting instead focuses on introducing an event after the subject leaves view and inferring its consequences when the subject is revisited, combining mid-rollout instruction following \citep{yang2025longlive}, long-range entity consistency, and reasoning over the realized visual history.

\begin{figure*}[t]
\centering
\includegraphics[width=\textwidth]{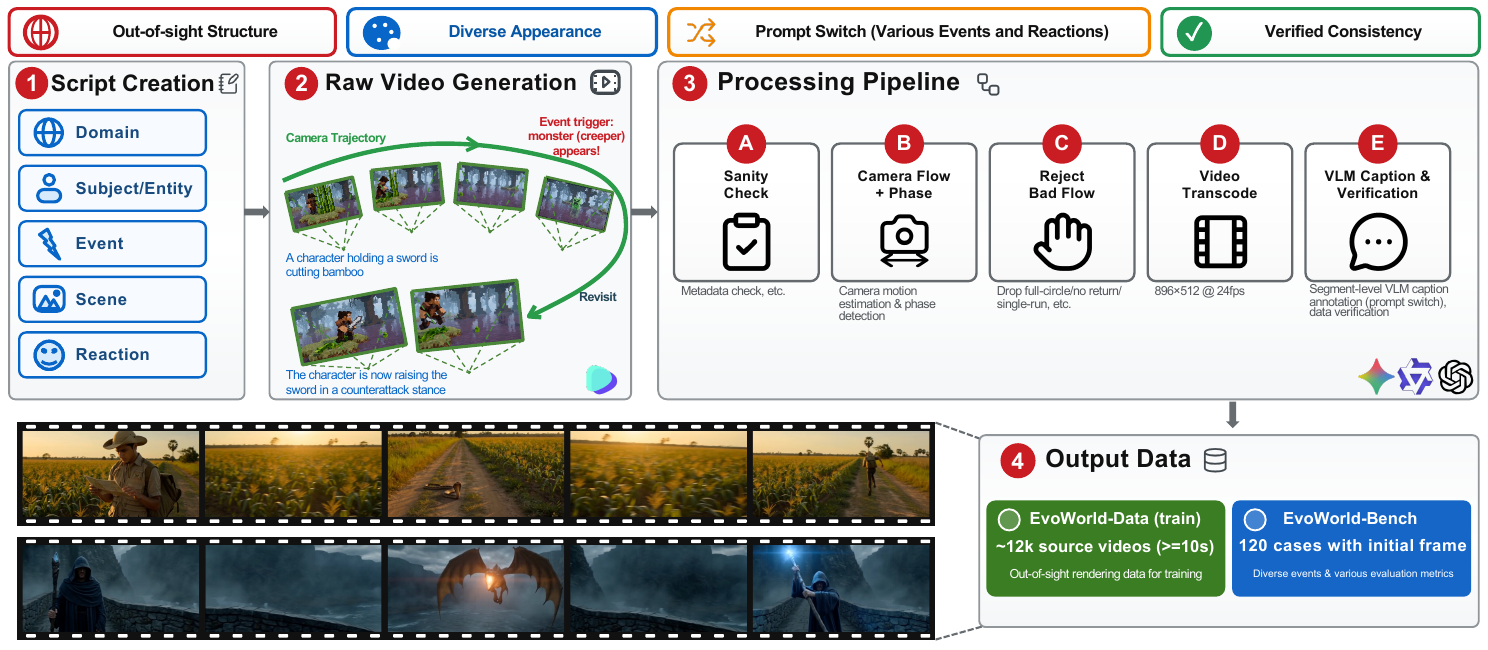}
\caption{Overview of the data engine for EvoWorld. Structured scenarios are rendered into single-take out-of-sight interactions, verified against visual evidence, temporally annotated, and converted into a training supervision dataset and evaluation benchmark.}
\label{data_engine}
\vspace{-0.3cm}
\end{figure*}

\section{\ours: Data and Method}\label{sec:evoworld}
\vspace{-0.2cm}
\subsection{Problem \& Task Setting}\label{sec:setting}\label{sec:setting-def}
\vspace{-0.2cm}
A world evolves continuously rather than only where and when it is being observed. Its entities should therefore remain active participants in that evolution: as new events occur and environmental conditions change, they should autonomously interact with those changes and exhibit appropriate responses, including while they are outside the current field of view. Leaving the frame should not freeze an entity or remove it from the unfolding dynamics---it is \emph{out of sight, but not out of the world}. We therefore formulate \emph{Out-of-Sight Causal Evolution} as a task for measuring whether a video world model or generation system can maintain a coherent account of off-screen entities and the relevant world state, infer how intervening events should alter that state and affect those entities, and render causally plausible outcomes, e.g., the entities' responses, when the scene is revisited. Here, \emph{causal evolution} means that the revealed post-event state must follow plausibly from the event actually realized, rather than merely remain visually consistent with the preceding frames.

Specifically, as shown in \cref{fig:challenge_compare}, an \emph{Out-of-Sight Causal Evolution} rollout can be abstracted into at least three semantic parts (each corresponding to at least one prompt):
(1) $\mathcal{P}_{\mathrm{init}}$ (\emph{initial}): the part that covers the initial state of the scenario and the main subjects. Condition $p_{init}$ specifies the subjects and scene, and the camera remains static or begins to move away from the subject.
(2) $\mathcal{P}_{\mathrm{event}}$ (\emph{event}): the part that introduces an event after the initial state, where the subject has left the frame because the camera view has moved away. Condition $p_{event}$ introduces an event (e.g., ``something is on fire''), and the camera may remain static, continue moving away to show the event, or begin moving back toward the original region.
(3) $\mathcal{P}_{\mathrm{revisit}}$ (\emph{revisit}): the part in which the camera returns to the initial region and reveals the resulting post-event world state, including the reappearance of the subject and its causally plausible response to the event. The ground-truth condition $p_{revisit}$ describes this resulting state and response, whereas the benchmark provides a blind revisit instruction without prescribing the concrete outcome.
These parts describe semantic roles rather than a fixed number of generated segments. Each part may span one or more autoregressive chunks/segments. We use this three-part abstraction to isolate the minimal causal chain while allowing longer rollouts and additional intermediate observations in real-world scenarios.
A system may internally infer a detailed control condition from the generated history, as our \textit{brain} does instead of pre-specifying the eventual reaction in the initial condition. The evaluation targets the complete event-to-reaction causal chain:
(1) the switch from $p_{init}$ to $p_{event}$ actually instantiates the requested event;
(2) the camera leaves and later revisits the original subject and its corresponding region (not necessarily the exact initial frame or viewpoint, since the subject may have moved during the out-of-sight interval);
(3) at revisit, the subject preserves its identity and core visual appearance without severe drift, and the surrounding scene and background remain reasonably consistent;
(4) the revealed post-event world state is causally plausible given the realized event, including an appropriate change in the subject's state or behavior, while maintaining natural motion and visual quality.




\subsection{\ours-Data \& Bench Construction}\label{sec:data}

\cref{data_engine} shows the overall data construction pipeline of EvoWorld. We here introduce the main components of the pipeline. Additional construction details are provided in Appendix~\cref{app:data-details}.


\begin{figure*}[t]
\centering

\includegraphics[width=\textwidth]{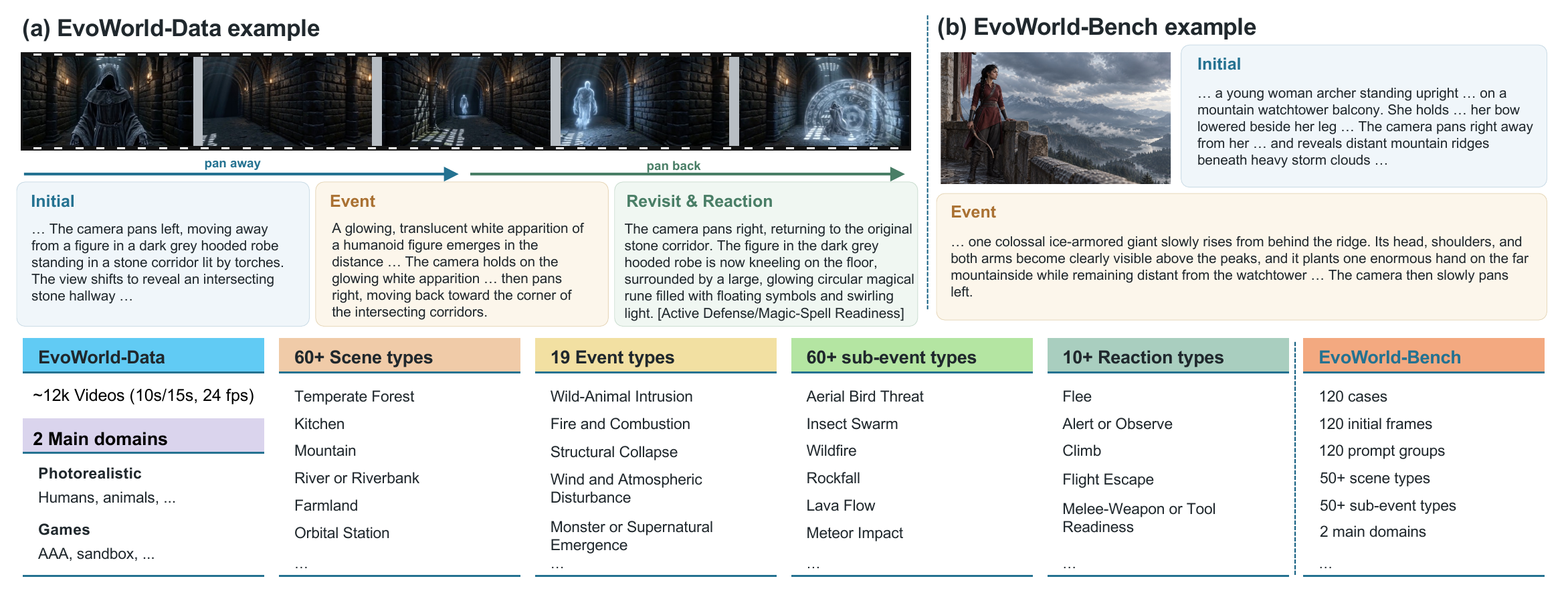}
\vspace{-0.7cm}
\caption{Examples and statistics of \ours-Data and \ours-Bench. The upper part illustrates representative training and benchmark examples, while the lower part summarizes their scale and scenario diversity.}
\label{fig:data-stats}
\vspace{-0.5cm}
\end{figure*}

\paragraph{EvoWorld-Data.}\label{sec:data-construction} (1) \textbf{Scenario design and video data generation.}
Existing real-world video datasets rarely contain samples that align with our experimental setting. We therefore construct a controlled generative dataset. The curated data aims to make event-conditioned world-state evolution observable, with the event-to-reaction causal chain separately auditable at each stage.
Scenarios are represented by structured tuples containing subjects and their appearance, scenes, events, compatible reactions, and camera movements. We compose tuples under semantic compatibility constraints---for example, the selected events must admit a visually observable response from the chosen subjects---and render them into a single-take generation script with five beats: initial scenario with the main subject $\rightarrow$ move the camera view away $\rightarrow$ show event $\rightarrow$ move the camera view back $\rightarrow$ revisit and reveal the resulting state and response. The full scripts are written by a VLM (Gemini3.1-Pro \citep{geminipro31}). Videos are then generated by the video foundation models \citep{seedance2026seedance2.0,seedance2.5} based on the designed script.
(2) \textbf{Verification and annotation.}\label{sec:data-verification}
Generated videos do not necessarily obey their scripts, so we verify visual evidence after generation. Specifically, we use Gemini3.1-Pro \citep{geminipro31} to check that each clip is a continuous shot, the designated subject remains visually consistent, the subject leaves and reappears, the event occurs in the intended view, the causal reaction is visible, and no severe hallucination invalidates the example. We further estimate global camera motion from Lucas--Kanade feature tracks using a RANSAC-based affine fit \citep{fischler1981random}.
To ensure the camera motion correctness, we reject full-circle camera motions, missing returns, single-direction motion, and unreliable direction estimates.
For segment-wise supervision, we partition each video into multiple intervals and describe them jointly so that adjacent captions form a coherent progression. We audit these captions for future-event spoilers, references to an absent subject, malformed camera instructions, and excessive adjacent-caption overlap.




\paragraph{EvoWorld-Bench.}\label{sec:data-benchmark}\label{sec:data-metrics}
(1) \textbf{Benchmark Construction}. We construct \ours-Bench as a 120-case evaluation set spanning photorealistic (69 cases) and game/virtual (51 cases) domains. Its initial frames have two complementary sources: 66 are extracted from held-out videos disjoint from the training set, while 54 are generated with GPT-Image-2 \citep{openai2026gptimage2}. Each case is organized into three semantic parts: an initial state, an event, and a revisit. After semantic deduplication, the benchmark covers 50+ fine-grained scene types and sub-event types, as summarized in \cref{fig:data-stats}.
Moreover, we conduct dedicated similarity screening against all training videos. For every benchmark case, we retrieve potentially similar training samples using both frame visual similarity and the lexical and embedding-based semantic similarity of the initial-event prompt pair. We then inspect retrieved candidate pairs with a VLM-based two-pass audit and human adjudication of the ambiguous cases. We remove any candidate judged highly similar, leaving no source-clip overlap, initial-event prompt-pair overlap, or human-confirmed high-similarity pair in the final benchmark.
(2) \textbf{Evaluation Metrics.}
We evaluate the complete rollout using Event Success, Revisit Success, Causal Plausibility, Reaction Quality, Visual Integrity, and revisit subject and background consistency.
Detailed inputs, scoring criteria, and aggregation rules are provided in Appendix~\cref{app:evoworld-bench}.

\subsection{Method}\label{sec:method}

\subsubsection{Overview}\label{sec:method-overview}

\begin{figure}[t]\centering
\includegraphics[width=\linewidth]{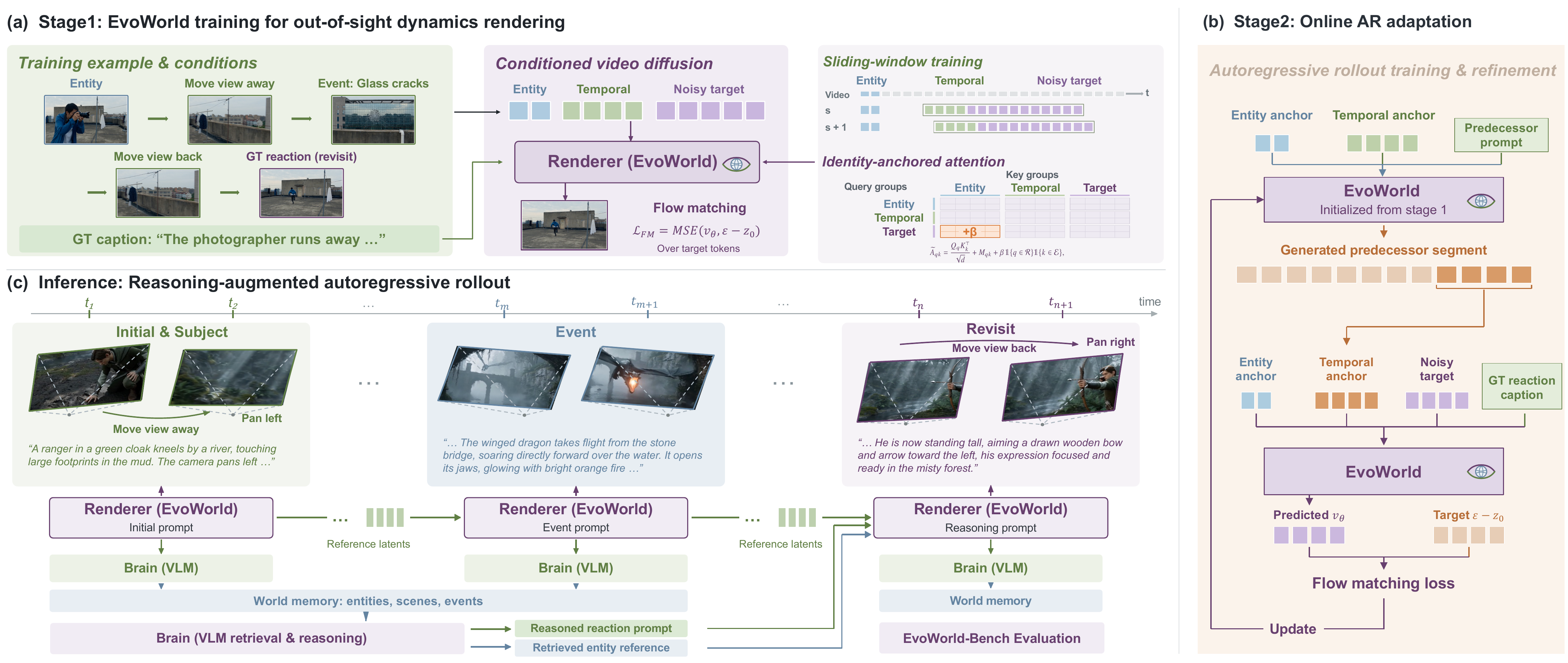}
\vspace{-0.3cm}
\caption{Overview of the \ours\ framework. (a) Stage 1 learns out-of-sight reactive rendering. (b) Stage 2 refines the renderer through online autoregressive adaptation. (c) Inference uses brain--renderer collaboration for \textit{Out-of-Sight Causal Evolution}.}\label{fig:method}
\vspace{-0.2cm}
\end{figure}

Most existing video world models provide little explicit support for interpreting what occurred during generation, maintaining a persistent account of entities outside the current view, or inferring how those entities should evolve before they reappear. We therefore introduce \ours, a framework for \textit{Out-of-Sight Causal Evolution} that explicitly separates \emph{world reasoning} from \emph{visual rendering} (\cref{fig:method}). A VLM \emph{brain} observes every generated segment and records the realized entities, scene, and events. After an event occurs, it reasons over this history and predicts how the relevant world state, including the off-screen subject, should have evolved. A segment-autoregressive \emph{visual renderer} then generates the next segment from the inferred instruction, retrieved entity evidence, and recent generated latents.
We equip the renderer with complementary entity and temporal anchors for long-range identity and short-range continuity. Identity-Anchored Attention (IAA) strengthens information flow from the entity anchor when the subject reappears. Our two-stage training first learns subject preservation and reactive rendering from clean ground-truth windows, and then performs online autoregressive rollout training from self-generated predecessor context. This Self-Forcing-style adaptation \citep{huang2026self_forcing} narrows the train-inference gap and refines model performance.

\subsubsection{Entity and temporal anchors}\label{sec:method-anchors}

The entity anchor carries subject appearance across a long out-of-view interval. During training, it is selected from ground-truth frames in which the target subject is clearly visible. At inference time, the VLM inspects sampled frames from the generated history, identifies a clear and unoccluded observation of the target subject, and retrieves the corresponding latents. This helps the renderer maintain subject identity and appearance when the camera revisits.
For temporal anchor, we adopt the four latents immediately preceding the target window, which participate in within-window full attention. This anchor is selected deterministically since recency is precisely the desired signal for view and motion continuity. Stage 1 takes these latents from the ground-truth videos, whereas stage 2 replaces them with the tail of the autoregressive rollout. At inference time they are always taken from the latest generated segment. The two anchors are therefore complementary: entity retrieval spans the long-term identity gap, while the temporal anchor bridges the local autoregressive boundary.

\subsubsection{Identity-Anchored Attention}\label{sec:method-iaa}

Although an appended entity reference participates in self-attention, it may not receive sufficient attention when the subject reappears: its contribution can be diluted by the much larger target and temporal context. We therefore introduce Identity-Anchored Attention (IAA), which makes the retrieved entity tokens a soft attention sink only for revisit samples. Let $\mathcal{R}$ denote target-query tokens in a revisit window and $\mathcal{E}$ the key tokens belonging to the entity anchor. For an attention head, IAA changes the pre-softmax score from $A$ to
\begin{equation}
\widetilde A_{qk}
= \frac{Q_q K_k^\top}{\sqrt{d}} + M_{qk}
+ \beta\,\mathds{1}\{q\!\in\!\mathcal{R}\}\mathds{1}\{k\!\in\!\mathcal{E}\},
\label{eq:iaa}
\end{equation}
where $M_{qk}$ is the base additive self-attention mask, $\beta>0$ controls identity preference, and $\mathds{1}\{\cdot\}$ denotes the indicator function. 
In full-attention video sequence, $M_{qk}=0$ for valid token pairs and takes a large negative value only for masked or padded pairs.
The IAA bias is directional: revisit tokens read from the entity anchor, while unrelated query-key pairs are unchanged. In implementation, we multiply the mask entries corresponding to the target-query--entity-key block by $e^\beta$ before LTX converts the mask to log space, which is exactly equivalent to adding $+\beta$ in \cref{eq:iaa}. IAA is enabled only for samples marked as revisit during training and for the revisit chunk during inference.

\begin{table*}[t]
\centering
\caption{Comparison results (\%) on the \ours-Bench. Best results are shown in bold and second-best results are underlined. \emph{Core} denotes the proportion of cases concurrently achieving successful revisit, event realization, and full causal plausibility.
}
\vspace{-0.2cm}
\label{tab:main}
\small
\setlength{\tabcolsep}{3.5pt}
\renewcommand{\arraystretch}{0.94}
\resizebox{\textwidth}{!}{%
\begin{tabular}{l*{9}{>{\centering\arraybackslash}m{0.10\textwidth}}}
\toprule
& \multicolumn{5}{c}{Task metrics} & \multicolumn{4}{c}{Visual quality, consistency, and motion} \\
\cmidrule(lr){2-6}\cmidrule(lr){7-10}
Method & \mbox{Core} & \mbox{Revisit} & \mbox{Event} & \mbox{Causal} & \mbox{React.} & \mbox{Visual} & \mbox{Subj. Cons.} & \mbox{Bg. Cons.} & \mbox{Motion} \\
\midrule
LingBot-World \citep{lingbot-world-v1}       & 15.8 & 92.5 & 77.5 & 29.2 & 31.3 & 68.8 & 87.6 & 94.6 & 98.0 \\
LingBot-World-v2  \citep{lingbot-world-v2}      & 20.0 & 76.7 & 68.3 & 27.1 & 25.4 & 76.7 & 82.6 & 92.5 & 98.4 \\
YUME-1.5 \citep{mao2026yume1.5}                 & 15.8 & 62.5 & 86.7 & 22.5 & 20.0 & 65.4 & 72.9 & 88.7 & 99.0 \\
SANA-WM \citep{zhu2026sana-wm}                  & 13.3 & 88.3 & 67.5 & 26.7 & 24.2 & 53.8 & 67.7 & 90.8 & 98.7 \\
Echo-WM \citep{zhang2026echowm}                 & 16.7 & 60.8 & 66.7 & 22.5 & 21.7 & \underline{82.9} & \underline{97.6} & 89.2 & 98.7 \\
Evoke  \citep{yin2026alaya-evoke}               & 20.0 & 95.8 & 80.0 & \underline{32.9} & \underline{36.3} & 78.3 & 93.5 & 95.4 & 99.0 \\
Evoke-Turbo \citep{yin2026alaya-evoke}          & \underline{21.7} & \underline{98.3} & \underline{89.2} & 32.5 & 29.6 & 65.4 & 97.0 & \underline{96.2} & \underline{99.1} \\
\midrule
\rowcolor{oursrow}
\textbf{\ours\ (ours)} & \textbf{90.8} & \textbf{100.0} & \textbf{97.5} & \textbf{94.2} & \textbf{85.0} & \textbf{83.8} & \textbf{99.2} & \textbf{98.3} & \textbf{99.5} \\
\bottomrule
\end{tabular}
}
\vspace{-0.3cm}
\end{table*}

\subsubsection{Two-stage autoregressive training}\label{sec:method-training-rules}

\paragraph{Stage 1: Learning causally grounded reactive rendering with clean anchors.}
For each sliding training window, the loader supplies both the entity anchor and the temporal anchor as conditioning. The training sequence is organized as $[\text{entity anchor}\mid\text{temporal anchor}\mid\text{target}]$. Both anchors remain clean at timestep zero and carry zero loss weight; only the target latents are noised and supervised by the flow-matching objective. IAA is activated on revisit windows. This stage teaches the renderer core capabilities such as event execution, event-conditioned reactions, and the roles of long-range identity and short-range temporal conditioning.

Formally, let $\mathbf{e}$, $\mathbf{c}$, and $\mathbf{z}$ denote the clean entity-anchor, temporal-anchor, and target latents, respectively; let $p$ be the target prompt and $r\in\{0,1\}$ the revisit indicator. For $\epsilon\sim\mathcal{N}(0,I)$ and a sampled noise level $\sigma$, we construct
\begin{equation}
\mathbf{x}^{\sigma}
=\left[\mathbf{e}\mid\mathbf{c}\mid
(1-\sigma)\mathbf{z}+\sigma\epsilon\right],
\qquad
\mathbf{s}^{\sigma}
=\left[\mathbf{0}_{|\mathbf e|}\mid\mathbf{0}_{|\mathbf c|}\mid
\sigma\mathbf{1}_{|\mathbf z|}\right],
\label{eq:stage1-input}
\end{equation}
where $\mathbf{s}^{\sigma}$ is the per-token timestep input, and $\mathbf{0}_n$ and $\mathbf{1}_n$ denote length-$n$ all-zero and all-one vectors, respectively. Thus the two anchors remain clean at timestep zero, whereas every target token receives timestep $\sigma$ and follows the flow-matching interpolation. The renderer $v_\theta$ predicts the target velocity $\epsilon-\mathbf{z}$, and stage 1 training minimizes
\begin{equation}
\mathcal{L}_{\mathrm{S1}}
=\mathbb{E}_{(\mathbf e,\mathbf c,\mathbf z,p,r),\,\sigma,\epsilon}
\left[
\frac{1}{|\Omega|D}
\sum_{i\in\Omega}
\left\|
v_\theta(\mathbf{x}^{\sigma},\mathbf{s}^{\sigma},p,r)_i
- (\epsilon_i-\mathbf{z}_i)
\right\|_2^2
\right],
\label{eq:stage1-fm}
\end{equation}
where $\Omega$ contains target tokens and $D$ is the latent channel dimension. The indicator $r$ controls whether the IAA bias in \cref{eq:iaa} is enabled.

\paragraph{Stage 2: Online autoregressive adaptation.}
Stage 1 is trained with a clean ground-truth temporal anchor, but deployment conditions each new segment on latents produced by the model itself. Small errors in appearance, motion, and geometry can therefore accumulate at precisely the event-to-revisit transition. We reduce this exposure gap with a lightweight second stage initialized from stage 1. For every selected event-to-revisit pair, the currently updated renderer performs an online one-hop rollout of the immediately preceding segment. This predecessor rollout is also conditioned on a clean entity anchor, its ground-truth temporal anchor, and the predecessor's conditioning prompt.
We take the final four latents of this self-generated predecessor as the temporal anchor $\widetilde{\mathbf{c}}_t$ for the revisit target, replacing its ground-truth temporal anchor. The resulting Stage-2 input is
\begin{equation}
\widetilde{\mathbf{x}}_t^{\sigma}
=\left[\mathbf e_t\mid\widetilde{\mathbf c}_t\mid
(1-\sigma)\mathbf z_t+\sigma\epsilon\right],
\end{equation}
with the per-token timestep input $\mathbf{s}_t^{\sigma}$ defined as in \cref{eq:stage1-input}. We optimize the same target-only velocity-matching objective as in Stage 1, now conditioned on the self-generated temporal anchor:
\begin{equation}
\mathcal{L}_{\mathrm{S2}}
=\mathbb{E}_{t,\sigma,\epsilon}
\left[
\frac{1}{|\Omega_t|D}
\sum_{i\in\Omega_t}
\left\|
v_\theta(\widetilde{\mathbf{x}}_t^{\sigma},
\mathbf{s}_t^{\sigma},p_t,r_t)_i
-(\epsilon_i-\mathbf z_{t,i})
\right\|_2^2
\right],
\label{eq:stage2-ar}
\end{equation}
where $\Omega_t$ contains only target tokens, $p_t$ is the target prompt, and $r_t$ is the revisit indicator controlling IAA. The entity and temporal anchors remain loss-free conditioning. 
Additional training details for the two-stage construction are provided in Appendix~\cref{app:training-algorithm}.

\subsubsection{VLM Brain: Perception and Reasoning}\label{sec:method-inference-vlm}

At inference time, our framework adopts a VLM as \textit{brain} to perceive and reason over the world evolution continuously. After every completed segment, the renderer decodes sampled frames for the \textit{brain}, which identifies and records the visible entities, appearances, object states, scenes, and any realized changes. These observations accumulate into a persistent world memory. Once an injected event has been rendered and observed, the \textit{brain} records what actually occurred, retrieves the identity evidence, and reasons over the accumulated context to infer a causally plausible evolution of the relevant world state, including how the off-screen subject should respond. It then expresses the inferred state as a concrete generation prompt $\hat p$ for the relevant subsequent segment. Additional details are provided in Appendix~\cref{app:vlm-reasoning}.

\section{Experiments}\label{sec:exp}
\vspace{-0.2cm}
\subsection{Experiment Setup}\label{sec:exp-setup}

We compare \ours\ with recent interactive video world models that support autoregressive generation: LingBot-World \citep{lingbot-world-v1}, LingBot-World-v2 \citep{lingbot-world-v2}, YUME-1.5 \citep{mao2026yume1.5}, SANA-WM \citep{zhu2026sana-wm}, Echo-WM \citep{zhang2026echowm}, and the standard and Turbo variants of Evoke \citep{yin2026alaya-evoke}. All baselines receive the same initial, event, and blind revisit prompts without a prescribed reaction, switched at matched semantic boundaries through their native autoregressive interfaces. Implementation details and additional general video models with one-shot results under a different, composed-prompt setting are provided in Appendix~\cref{app:baselines}.
We evaluate Out-of-Sight Causal Evolution on EvoWorld-Bench with the following metrics: \emph{Event Success}, \emph{Revisit Success}, \emph{Causal Plausibility}, \emph{Reaction Quality}, and \emph{Visual Integrity}. We additionally assess subject and background consistency between the opening and revisited views while allowing causally appropriate changes in pose, state, and viewpoint, and report \emph{Motion Smoothness} using VBench \citep{huang2024vbench}. Our primary task-level measure, \emph{Core Task Success}, requires Event Success, Revisit Success, and full Causal Plausibility (i.e., causal plausibility score=1) to hold simultaneously. Appendix~\cref{app:evoworld-bench} provides the complete evaluation protocol.

\begin{figure*}[t]
\centering
\begin{minipage}[t]{0.475\textwidth}
\vspace{0pt}
\centering
\captionof{table}{Component ablation of \ours. Cumulative effects of IAA and Stage-2 online rollout training on \ours-Bench. All values are percentages, and column-wise best results are bold.}
\label{tab:component-ablation}
\scriptsize
\setlength{\tabcolsep}{3.2pt}
\renewcommand{\arraystretch}{1.18}
\resizebox{\linewidth}{!}{%
\begin{tabular}{lccccc}
\toprule
Setting & Core & Revisit & Event & Causal & React. \\
\midrule
Stage 1 w/o IAA       & 82.5 & \textbf{100.0} & 96.7 & 88.8 & 80.8 \\
Stage 1 + IAA         & 84.2 & 99.2 & \textbf{98.3} & 88.7 & 83.9 \\
\rowcolor{oursrow}
Stage 1 + Stage 2      & \textbf{90.8} & \textbf{100.0} & 97.5 & \textbf{94.2} & \textbf{85.0} \\
\bottomrule
\end{tabular}%
}
\end{minipage}
\hfill
\begin{minipage}[t]{0.505\textwidth}
\vspace{-0.5cm}
\centering
\includegraphics[width=\linewidth]{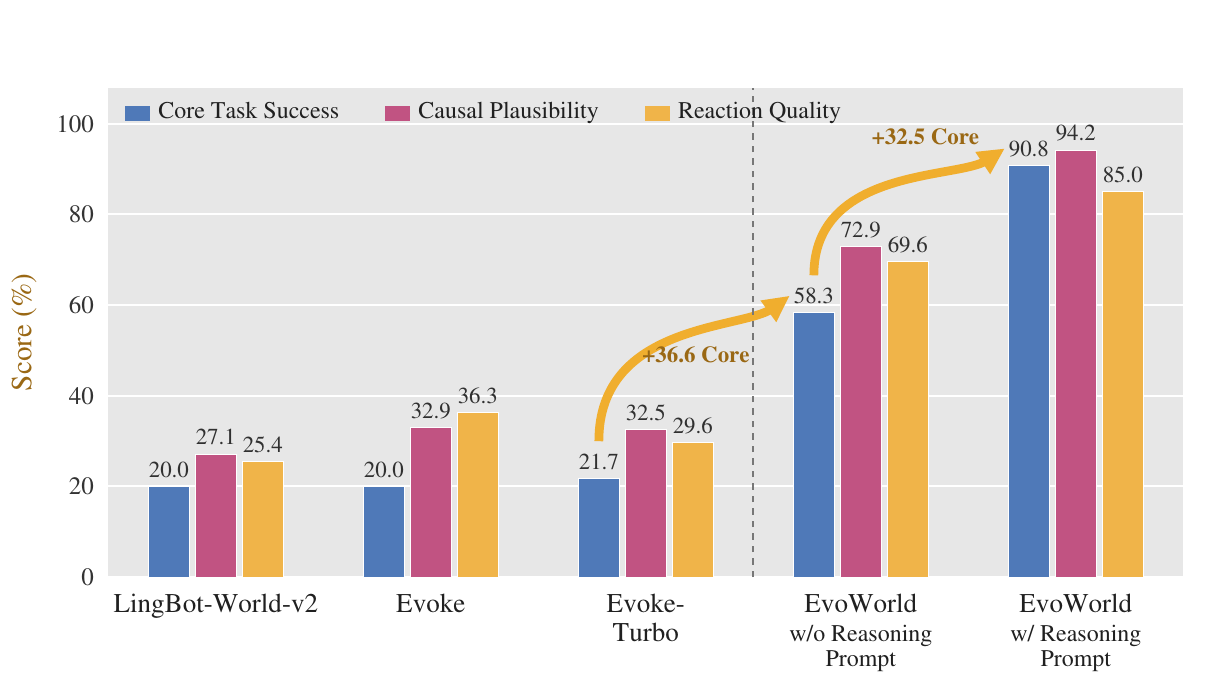}
\vspace{-0.8cm}
\captionof{figure}{Effect of event-conditioned reasoning prompts on Out-of-Sight Causal Evolution.}
\label{fig:brain-ablation}
\end{minipage}
\vspace{-0.2cm}
\end{figure*}

\subsection{Implementation Details}\label{sec:exp-impl}
\vspace{-0.1cm}
The VLM brain is instantiated with Qwen3.8-27B \citep{yang2025qwen3}, served through vLLM \citep{vllm}, while the visual renderer is initialized from LTX-2.5 \citep{hacohen2026ltx-2}. In Stage 1, we train a rank-128 LoRA for 8,000 steps with a learning rate of $1\!\times\!10^{-4}$, using a per-GPU batch size of 1 on eight GPUs and eight gradient-accumulation steps, resulting in an effective batch size of 64. We set the IAA bias to $\beta=3.0$. In Stage 2, we initialize from the Stage-1 checkpoint and conduct online autoregressive rollout training for 200 steps at $2\!\times\!10^{-5}$.
Inference uses the official distilled LoRA and eight-step sampler to generate videos at $896\!\times\!512$ resolution and 24 fps. Complete training and inference settings are provided in Appendix~\cref{app:hparams,app:training-algorithm}.

\vspace{-0.1cm}
\subsection{Results of Out-of-Sight Causal Evolution}\label{sec:exp-main}

\textbf{Out-of-sight causal evolution.}
As shown in \cref{tab:main}, \ours\ achieves 90.8\% Core Task Success, exceeding the second-best baseline by 69.1 percentage points. Existing world models can often execute individual parts of the rollout: for example, Evoke-Turbo reaches 98.3\% Revisit Success and 89.2\% Event Success. However, its Core Task Success and Causal Plausibility results reveal that successful camera control and event rendering do not by themselves produce a coherent event--reaction chain. In contrast, \ours\ obtains promising Revisit Success, Event Success, and Causal Plausibility. Its Causal Plausibility and Reaction Quality outperform other methods by a significant margin, demonstrating a substantially stronger ability to render causally plausible interaction evolution in the out-of-sight setting.
\textbf{Consistency and generation quality.}
The improvement in task success does not come at the cost of appearance preservation or visual quality. \ours\ achieves promising Subject Consistency, Background Consistency, Visual Integrity, and Motion Smoothness. Its consistently high VLM-based scores and VBench \citep{huang2024vbench} motion evaluation indicate strong \textit{Out-of-Sight Causal Evolution} performance while preserving generation quality.

\begin{figure*}[t]
\centering
\begin{minipage}[t]{0.35\textwidth}
\vspace{0pt}
\centering
\captionof{table}{Transfer of the VLM-brain reasoning prompt (\%).}
\vspace{-0.3cm}
\label{tab:brain-transfer}
\scriptsize
\setlength{\tabcolsep}{4pt}
\renewcommand{\arraystretch}{1.05}
\resizebox{0.90\linewidth}{!}{%
\begin{tabular}{lccc}
\toprule
Method & Core & Causal & React. \\
\midrule
Evoke-Turbo & 21.7 & 32.5 & 29.6 \\
+ Brain prompt & \underline{45.0} & \underline{53.3} & \underline{50.8} \\
\rowcolor{oursrow}
\textbf{\ours\ (ours)} & \textbf{90.8} & \textbf{94.2} & \textbf{85.0} \\
\bottomrule
\end{tabular}
}

\vspace{-0.1cm}

\captionof{table}{Event--prompt alignment based on CLIP score. }
\label{tab:event-prompt-alignment}
\scriptsize
\setlength{\tabcolsep}{4.5pt}
\renewcommand{\arraystretch}{1.05}
\resizebox{0.9\linewidth}{!}{%
\begin{tabular}{lcc}
\toprule
Method & EPA-Abs & EPA-TopK \\
\midrule
LingBot-World & 0.2716 & 0.2827 \\
LingBot-World-v2 & 0.2740 & 0.2860 \\
SANA-WM           & 0.2789 & 0.2872 \\
Echo-WM           & 0.2769 & 0.2896 \\
Evoke             & \underline{0.2893} & \underline{0.3043} \\
Evoke-Turbo       & 0.2885 & 0.3025 \\
\rowcolor{oursrow}
\textbf{\ours\ (ours)} & \textbf{0.2942} & \textbf{0.3126} \\
\bottomrule
\end{tabular}
}
\end{minipage}
\hfill
\begin{minipage}[t]{0.62\textwidth}
\vspace{0pt}
\centering
\includegraphics[width=\linewidth]{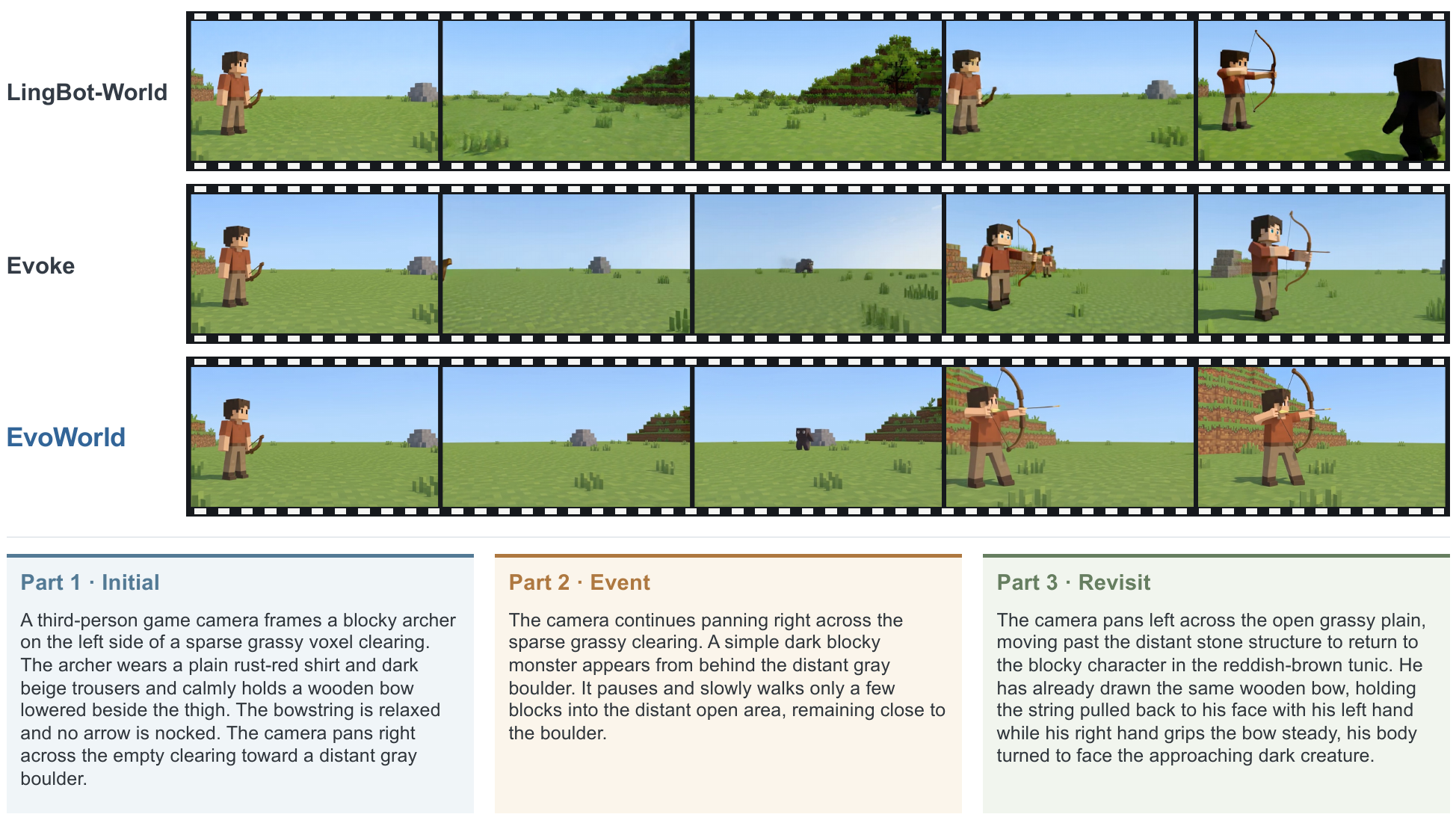}
\vspace{-0.5cm}
\captionof{figure}{Qualitative comparison under the same VLM-brain reasoning prompt. 
}
\label{fig:bow-case}
\end{minipage}
\end{figure*}

\vspace{-0.2cm}
\subsection{Ablation Study and Additional Analysis}\label{sec:exp-ablation}

In this section, we adopt a research-question-driven analysis to analyze and validate the effectiveness of our method, including ablation studies and other analyses.
\vspace{-0.3cm}
\paragraph{Do IAA and Stage-2 online rollout training improve out-of-sight causal evolution?}
As shown in \cref{tab:component-ablation}, IAA increases Core Task Success from 82.5\% to 84.2\% and Reaction Quality from 80.8\% to 83.9\%, demonstrating the benefit of strengthening attention to retrieved entity evidence. Stage 2 further raises Core Task Success to 90.8\% and Causal Plausibility to 94.2\%. These results indicate that both IAA and stage-2 online rollout training improve the model's capability for \textit{out-of-sight causal evolution}.
\vspace{-0.3cm}
\paragraph{How do event-conditioned reasoning prompts affect different visual renderers?}
We replace the brain-generated reasoning prompt with the blind revisit prompt while retaining the trained \ours\ renderer and visual conditioning. As shown in \cref{fig:brain-ablation}, performance drops substantially without event-conditioned reasoning, confirming its importance for inferring an appropriate response from the realized event. This variant nevertheless remains stronger than the baselines, indicating that the proposed data and renderer training already provide a strong event-conditioned causal-response prior. Moreover, applying the same reasoning prompts to Evoke-Turbo consistently improves its task metrics (\cref{tab:brain-transfer}), demonstrating that the benefit transfers across visual renderers.

\vspace{-0.3cm}

\paragraph{How do models compare in CLIP-based event--prompt alignment?}
As an additional reference to the VLM-based Event Success metric, we compare the generated event segment with its event prompt using CLIP ViT-B/32 \citep{clip}. We recover each model's actual event-prompt interval, uniformly sample eight frames, and report their mean similarity (EPA-Abs) and the mean of the two highest similarities (EPA-TopK). As shown in \cref{tab:event-prompt-alignment}, \ours\ obtains the highest scores. These results should be interpreted only as a relative comparison: event prompts contain substantial shared scene and camera descriptions, while CLIP similarity cannot precisely determine whether a fine-grained event action has been correctly executed.
\vspace{-0.3cm}
\paragraph{Does a shared reasoning prompt alone ensure faithful causal evolution?}
We provide LingBot-World, Evoke, and \ours\ with the same VLM-inferred revisit prompt to isolate renderer capability. As shown in \cref{fig:bow-case}, the baselines exhibit hallucinated event entities, subject duplication, or coarse revisit frames, whereas \ours\ preserves the subject and scene while rendering a coherent defensive response. Thus, an appropriate reasoning prompt must be paired with a renderer capable of preserving history and realizing the inferred evolution.

\section{Conclusion}\label{sec:conclusion}
\vspace{-0.3cm}
In this paper, 
we formulate \emph{Out-of-Sight Causal Evolution} as a concrete setting for investigating autonomous world evolution and construct \ours-Data/Bench to support learning and evaluating event realization, causally plausible post-event states, revisit consistency, and visual quality. We further present the \ours framework, a reasoning-augmented brain-renderer collaboration framework that continuously perceives and renders the world dynamics. The renderer is trained with identity-anchored attention and a two-stage procedure that combines supervised learning of reactive rendering with autoregressive rollout training. Experiments show that \ours\ substantially improves \textit{Out-of-Sight Causal Evolution} capabilities over recent video world models while preserving generation quality and consistency.
We hope this study can provide a reference for future research in reasoning-enhanced world evolution. Limitations and future directions are discussed in Appendix~\cref{app:limitations}.

\bibliography{evoworld_arxiv}
\bibliographystyle{evoworld_arxiv}

\newpage
\appendix
\section*{Appendix}

\section{EvoWorld-Data and EvoWorld-Bench Details}\label{app:data-details}\label{app:evoworld-bench}

\subsection{EvoWorld-Data construction}

\paragraph{Structured scenario generation.}
Each training example begins from a structured tuple specifying the domain, scene, subject and appearance, event, compatible reaction, and camera trajectory. Compatibility constraints exclude combinations whose consequence would be visually ambiguous or implausible for the designated subject. Gemini3.1-Pro expands the tuple into a continuous-shot script covering the initial observation, camera move-away, event reveal, camera move-back, and reacted subject. We render these scripts with Seedance~2.0 and 2.5 \citep{seedance2026seedance2.0,seedance2.5}. The retained training set contains around 12k videos, each at least ten seconds long.

\paragraph{Verification and temporal annotation.}
Generated candidates are checked for continuous-shot structure, subject identity, complete subject exit and re-entry, visible event realization, visible reaction, and severe rendering failures. Camera direction is verified independently from sparse Lucas--Kanade feature tracks \citep{lucas1981iterative} and a RANSAC affine fit \citep{fischler1981random}; clips with a missing reversal, full-circle motion, only unidirectional motion, or unreliable estimates are rejected. For each retained video, we record the subject-disappearance and reappearance times, camera phase boundaries, event interval, reaction interval, and generation provenance. Segment captions are produced jointly across adjacent intervals and audited to remove future-event spoilers, references to an absent subject, malformed camera commands, and redundant neighboring descriptions.

\subsection{Reviewed semantic taxonomy}
Raw generation labels contain aliases and style prefixes and therefore overestimate semantic diversity. We consolidate the \ours-Data labels into a two-level taxonomy: a coarse family and a fine-grained semantic type. Synonyms are merged, and rendering styles such as AAA, voxel, fantasy, or photorealistic do not create a new type unless the underlying event, environment, or reaction is semantically distinct. For \ours-Bench, we report only event and scene coverage: the benchmark does not define or expose target reaction categories, and any causally plausible response is accepted. \cref{tab:taxonomy-stats} reports the resulting human-reviewed statistics.

\begin{table}[h]
\centering
\caption{Reviewed semantic coverage of \ours-Data and \ours-Bench. Reaction categories are reported only for the training data; \ours-Bench does not define target reactions.}
\label{tab:taxonomy-stats}
\small
\setlength{\tabcolsep}{4.5pt}
\begin{tabular}{lrrrrrrr}
\toprule
Split & Videos & \multicolumn{2}{c}{Events} & \multicolumn{2}{c}{Scenes} & \multicolumn{2}{c}{Reactions} \\
\cmidrule(lr){3-4}\cmidrule(lr){5-6}\cmidrule(lr){7-8}
& / cases & Fam. & Types & Fam. & Types & Fam. & Types \\
\midrule
\ours-Data & 12k & 19 & 65 & 16 & 63 & 5 & 14 \\
\ours-Bench & 120 & 17 & 55 & 16 & 56 & -- & -- \\
\bottomrule
\end{tabular}
\end{table}

The event taxonomy includes, for example, fire and combustion, structural collapse, wild-animal intrusion, supernatural emergence, flooding, atmospheric disturbance, and vehicle failure. Reaction families in \ours-Data include evasive relocation and active defense, among others.

\subsection{EvoWorld-Bench construction and train--test screening}\label{app:bench-screening}
The 120 benchmark cases comprise 69 photorealistic and 51 game/virtual scenarios. Sixty-six initial frames come from held-out source videos and 54 are generated with GPT-Image-2 \citep{openai2026gptimage2}. We first remove exact source overlap and then screen every benchmark case against the complete training split at both the visual and semantic levels. Visual retrieval uses the benchmark initial frame to obtain its nearest training frames, while semantic retrieval operates on the paired initial and event prompts using lexical overlap and text-embedding similarity and requires both parts to come from the same training video. The union of these candidates is passed to a first VLM comparison. Pairs with substantial similarity in both the initial state and event are re-evaluated by a second multimodal pass that jointly receives the two initial frames and all four prompt fields. Ambiguous cases are manually adjudicated, and any benchmark candidate with a confirmed identical or near-identical training counterpart is removed or revised. A high first-frame similarity alone, or membership in the same scene or event category, is not treated as evidence of duplication.

\subsection{Prompt protocol and reaction-intensity calibration}\label{app:bench-prompts}
We label every event by its required reaction intensity. A \emph{strong} event requires an urgent response such as fleeing or taking cover, whereas a \emph{weak} event permits either a mild or strong response. 
This calibration prevents a visually present but implausibly weak response to a dangerous event from being counted as correct.

At evaluation time, models receive the initial prompt, the event prompt at the appropriate switch point, and a blind revisit instruction that asks the subject to react without specifying how. The model must therefore infer an appropriate response from the event realized during its own rollout; the benchmark neither exposes nor requires a predefined reaction type, and any causally plausible response is considered valid. We audit all three prompts to ensure that the initial prompt does not reveal the event, the event prompt does not reveal the eventual response, and the blind revisit prompt contains neither a specific reaction nor future visual details.

\subsection{Evaluation protocol}\label{app:bench-evaluation}
Our evaluation combines complementary sources of evidence rather than relying on a single judge. A strong VLM Qwen3.8 \citep{yang2025qwen3} assesses task execution, causal plausibility, visual integrity, and revisit consistency. Established tool-based metrics provide independent diagnostics: VBench \citep{huang2024vbench} measures motion smoothness, while CLIP ViT-B/32 \citep{clip} supplies the auxiliary event--prompt alignment comparison in \cref{tab:event-prompt-alignment}.

For the VLM-based metrics, we temporally stratify each generated rollout into its beginning, middle, and ending portions and uniformly sample 30 frames across the three portions. The judge receives these frames, the initial and event prompts, and the strong/weak event-intensity label used to calibrate reaction strength, but not a target reaction or ground-truth revisit description. It jointly inspects whether the camera returns, whether the requested event is realized, what response the subject exhibits, whether that response is causally plausible at the required intensity, and whether the revisited subject and environment remain consistent. Event and revisit success are binary; the remaining perceptual and behavioral criteria use the three-level scale $\{0,0.5,1\}$.

\paragraph{Event and revisit success.}
\emph{Event Success} is 1 only if the event described by the event prompt visibly occurs in the middle portion of the generated rollout; otherwise it is 0. \emph{Revisit Success} is 1 only if the camera first leaves the original subject and region and subsequently returns to the same location with the subject visible again. A reaction may cause the subject to leave the frame after being revisited without invalidating the revisit. A cut to an unrelated scene, or a return without the original subject, receives 0.

\paragraph{Causal plausibility.}
This metric evaluates whether the subject's changed behavior or state is a plausible consequence of the event that was actually rendered. The judge may use changes relative to stable landmarks---for example, retreating behind a doorway, moving behind cover, or relocating to higher ground---as evidence of an off-screen response. A score of 0 indicates no reaction when one is required, or a response that contradicts the event's causal logic; 0.5 indicates a plausibly related but insufficiently strong response to a strong event; and 1 indicates a causally appropriate response at the expected intensity. We do not require one prescribed action: escape, cover, safe relocation, or defensive readiness may all receive full credit when appropriate. If either Event Success or Revisit Success is 0, Causal Plausibility is set to 0 because the generated video provides no valid event--reaction chain to assess.

\paragraph{Reaction quality and visual integrity.}
\emph{Reaction Quality} isolates the visual realization of the response from its semantic appropriateness. It receives 0 for broken, distorted, or incoherent reaction motion; 0.5 for a visible but stiff or mildly artifacted response; and 1 for a natural and coherent action or changed state. If no changed reaction state is visible, the per-case value is null and is treated as 0 in the all-case mean reported for cross-model comparison. \emph{Visual Integrity} evaluates the complete rollout independently of task correctness: 0 denotes severe structural breakdown, hallucination, or melting; 0.5 denotes a watchable video with visible local artifacts, instability, or flicker; and 1 denotes a clean and coherent rollout without obvious breakdown, with particular attention paid to prompt-switch boundaries.

\paragraph{Revisit consistency.}
\emph{Subject Consistency} compares the opening subject with the revisited subject while explicitly allowing its pose, behavior, and state to change. A score of 0 denotes a different or severely drifted identity, 0.5 a recognizable subject with noticeable identity or appearance inconsistencies, and 1 a clearly preserved identity. If the subject does not reappear and therefore cannot be compared, the case receives a score of 0. The reported Subject Consistency is averaged over all 120 benchmark cases. \emph{Background Consistency} compares the opening and revisited locations while allowing modest viewpoint changes: 0 indicates a different scene, 0.5 the same general place with a major object or layout inconsistency, and 1 the same location with its principal structures and objects preserved. Minor viewpoint changes or incidental small-object differences do not reduce the score.

\paragraph{Motion smoothness.}
We report the Motion Smoothness dimension of VBench \citep{huang2024vbench} over all 120 generated videos. This metric is independent of the VLM judge and measures local temporal smoothness rather than completion of the event--revisit--reaction chain. Because it is highly saturated for modern video generators, we treat it as a secondary quality diagnostic rather than a primary task metric.

\paragraph{Core success and aggregation.}
Let $r_i$ and $e_i$ denote binary Revisit and Event Success for case $i$, and let $c_i\in\{0,0.5,1\}$ denote Causal Plausibility. We define
\begin{equation}
\mathrm{Core}=\frac{100}{N}\sum_{i=1}^{N}\mathds{1}\!\left[r_i=1\;\land\;e_i=1\;\land\;c_i=1\right].
\end{equation}
Thus Core Success requires all three task-critical conditions in the same case. Reaction Quality and Visual Integrity are reported separately and are intentionally excluded from Core, allowing task completion and rendering quality to be analyzed independently. Unless otherwise stated, three-level metrics are averaged over all 120 benchmark cases and reported as percentages.

\section{Training hyperparameters}\label{app:hparams}

\begin{table}[h]
\centering
\caption{Training and inference settings for \ours. Stage 2 inherits the Stage-1 LoRA architecture and IAA configuration.}
\label{tab:training-hparams}
\small
\setlength{\tabcolsep}{5pt}
\begin{tabular}{ll}
\toprule
Setting & Value \\
\midrule
Visual renderer & LTX-2.5 \citep{hacohen2026ltx-2} \\
VLM brain & Qwen3.8-27B \citep{yang2025qwen3} \\
Output resolution / frame rate & $896\!\times\!512$ / 24 fps \\
Entity-reference length (V2V train / I2V benchmark) & 2 / 1 temporal latent steps \\
Temporal-anchor length & 4 temporal latent steps \\
LoRA rank & 128 \\
IAA bias $\beta$ & 3.0 \\
\midrule
Stage-1 optimizer / precision & AdamW / bfloat16 \\
Stage-1 learning rate / steps & $1\!\times\!10^{-4}$ / 8,000 \\
Stage-1 effective batch size & 64 (8 GPUs, accumulation 8) \\
Stage-1 gradient clipping & 1.0 \\
\midrule
Stage-2 learning rate / steps & $2\!\times\!10^{-5}$ / 200 \\
Stage-2 online / GT-anchor ratio & 75\% / 25\% \\
Online and test-time sampler & official distilled adapter, 8 steps \\
\bottomrule
\end{tabular}
\end{table}

\paragraph{Training and benchmark inference modes.}
Both training stages optimize the renderer in a video-to-video (V2V) setting, using the first two temporal latent steps of the training clip as the entity reference. For benchmark evaluation, we use the same learned renderer directly in image-to-video (I2V) mode: the provided initial frame is encoded into one entity-reference latent step, after which generation proceeds autoregressively. Each later segment is conditioned on the final four latent steps of its generated predecessor. We also experimented with training the renderer directly in I2V mode and observed comparable benchmark performance; we therefore retain the unified V2V training configuration for the reported model. 

\section{Two-stage training procedure}\label{app:training-algorithm}

Stage 2 uses a \emph{one-hop} rollout rather than regenerating the entire history chain: it exposes the renderer to direct predecessor errors while keeping training tractable. Because the predecessor is regenerated by the current model as training progresses, the conditioning distribution co-evolves with the renderer rather than remaining fixed to a single offline snapshot. To prevent this focused adaptation from overwriting the initial, event, and clean-revisit behavior learned in Stage 1, we mix 25\% deterministic samples with ground-truth temporal anchors drawn across all phases and 75\% samples with online self-generated temporal anchors. At deployment, the resulting Stage-2 LoRA is used together with the same official distilled LoRA and eight-step sampler employed during online rollout.

\begin{algorithm}[h]
\caption{Two-stage anchor and online-rollout construction}\label{alg:train-select}
\begin{lstlisting}[basicstyle=\ttfamily\small]
def stage1_sample(clip, target):
    entity = opening_entity_anchor(clip)
    prefix = gt_predecessor_tail(target, count=4)
    return entity, prefix, target.is_revisit

def stage2_sample(clip, target, use_online):
    if not use_online:  # 25% Stage-1 rehearsal
        clip, target = sample_stage1_window()
        return stage1_sample(clip, target)
    assert target.is_revisit and prompt_changed(target)
    entity = opening_entity_anchor(clip)
    temporal_anchor = gt_temporal_anchor(target.prev, count=4)
    with no_grad():
        prev = rollout_one_hop(current_model,
                entity, temporal_anchor, target.prev.prompt,
                sampler="distilled-8step")
    prefix = stop_grad(prev.latents[-4:])
    return entity, prefix, True
\end{lstlisting}
\end{algorithm}

\section{VLM brain protocol}\label{app:vlm-reasoning}

The VLM brain is called after every generated segment rather than only at the final revisit. Each call receives temporally sampled frames from the newly generated segment together with the accumulated textual world record. Its first responsibility is observational: it records visible entities and their attributes, persistent scene landmarks, object states, camera movement, and any newly realized event. Before an event has been observed, the brain updates this record but does not propose a reaction. This separation prevents the initial observation from leaking future event or reaction information.

After the event segment, the brain reasons from the event that was actually rendered, not merely from the requested event prompt. It identifies the affected off-screen subject, infers one causally plausible changed state or behavior, and converts that inference into a concrete renderer prompt. The prompt specifies the camera return, preserved identity cues, the subject's post-event state, and visually realizable behavior, while avoiding hidden mental states or explanations that cannot be rendered. The benchmark remains open-ended: this inferred behavior need not match an exact reaction if it is causally plausible for the realized event.

For long-range identity conditioning, the framework can select a clear prior observation of the target entity and map it to the corresponding latent span. In the reported controlled I2V evaluation, the opening frame is deliberately constructed to show the target clearly and is used as the persistent one-step entity reference. Content-based retrieval is retained for longer or less constrained rollouts in which the clearest observation need not be the first frame. In either case, retrieval determines visual evidence, whereas the VLM-generated prompt determines the subsequent semantic state; the four most recent generated latent steps provide short-range temporal continuity independently of this decision.

\begin{figure*}[t]
\centering
\includegraphics[width=0.98\textwidth]{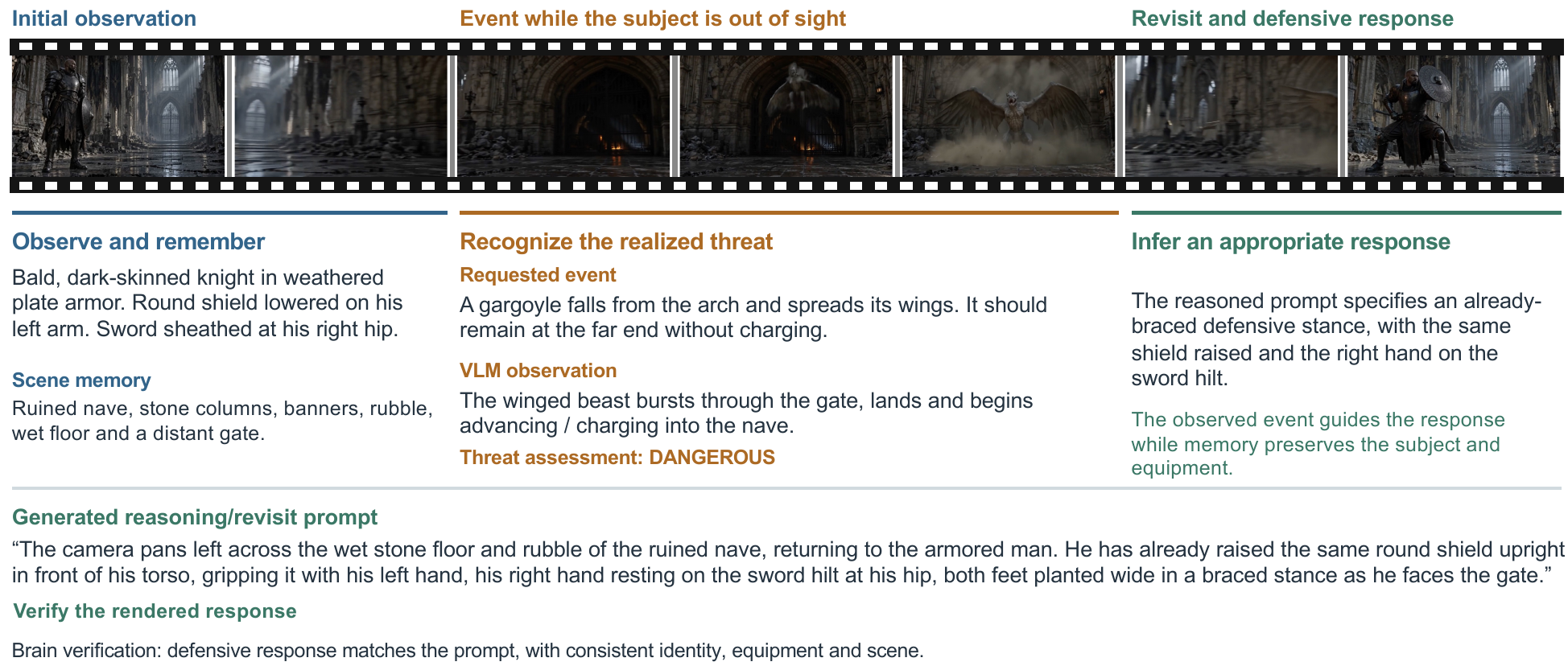}
\caption{Case study summarizing the VLM brain's observation, event-conditioned reasoning, prompt generation, and verification over an out-of-sight rollout.}
\label{fig:vlm-case-study}
\end{figure*}

\paragraph{Reasoning case study.}
\cref{fig:vlm-case-study} summarizes one reference brain-renderer interaction. The brain first records the knight's identity, equipment, and cathedral environment before the camera moves away. It then observes the event actually realized by the renderer: although the requested gargoyle was intended to remain near the gate, the generated creature begins advancing into the nave. Based on this observed threat and the accumulated subject memory, the brain produces a revisit prompt in which the knight has raised the same shield and adopted a braced defensive stance. The final observation verifies that the rendered response is causally appropriate while preserving the subject, equipment, and scene. This example illustrates how the brain conditions its reasoning on the realized visual history rather than relying only on the requested event prompt.

\section{Anchor layout and IAA implementation}\label{app:iaa-implementation}\label{app:rope-alignment}

Entity-reference and target-window tokens are combined through the native LTX conditioning path. Here, the target window denotes the complete generation-side latent sequence, including the clean temporal anchor and the target latents being denoised. Their physical order differs across code paths: training forms $[\text{entity reference}\mid\text{target window}]$, whereas inference appends the reference and forms $[\text{target window}\mid\text{entity reference}]$. IAA therefore derives the entity-key range from the actual reference length and layout instead of assuming a fixed prefix position. Runtime checks verify the reference length, the contiguous denoised-query span, and the target--reference boundary.

IAA is enabled only for revisit windows, which are oversampled by a factor of three in Stage 1. If an entity reference contains $n_e$ temporal latent steps and each step contains $H_\ell W_\ell$ spatial tokens, its key block contains $n_eH_\ell W_\ell$ tokens. For each denoised target query, we multiply the corresponding entity-key entries of the LTX attention mask by $\exp(\beta)$. LTX subsequently converts this mask to an additive log-space bias, producing the $+\beta$ term in \cref{eq:iaa}. We use $\beta=3.0$; other valid pairs are unchanged, and originally masked pairs remain masked. 

\section{Baseline reproduction details}\label{app:baselines}
\paragraph{Common prompt protocol.}
For every benchmark case, all autoregressive models receive the same three user-facing conditions: (i) an initial prompt describing the starting subject and scene, (ii) an event prompt introduced only after the camera has moved away, and (iii) a blind revisit prompt that asks the camera to return and the subject to react without revealing a particular reaction. The corresponding camera schedule moves the view away from the subject, exposes the event region, and then moves the view back. We inject each prompt at the nearest native chunk boundary and preserve the model's recurrent state, latent history, or memory cache across switches. Thus, no model is reset at the event or revisit boundary. In the main comparison, baselines directly execute the blind revisit prompt and receive neither VLM-generated reaction reasoning nor entity retrieval; \ours\ internally interprets the same blind request using its VLM brain. For a more detailed ablation, the prompt-transfer and shared-prompt analyses in \cref{tab:brain-transfer,fig:bow-case} also provide baseline renderers with the brain-generated reasoning prompt, allowing us to separate the contribution of event-conditioned reasoning from that of the proposed data and renderer training.

\paragraph{Compatibility of additional recent world models.}
We also conducted preliminary protocol-compatibility trials with HY-World~1.5 \citep{hyworld1.5}, Matrix-Game~3.0 \citep{wang2026matrixgame3}, and AlayaWorld \citep{team2026alayaworld}, but do not include them in the quantitative comparison because their released models did not reliably support the complete evaluation protocol. In particular, HY-World~1.5 frequently continued the preceding content while weakly realizing or entirely ignoring the event introduced by the mid-rollout prompt switch. Matrix-Game~3.0 and AlayaWorld likewise did not consistently complete the combined move-away, newly injected event, blind revisit, and event-conditioned reaction sequence under their available inference interfaces. This mismatch is unsurprising: these systems are trained primarily for their respective action-conditioned interaction, spatial persistence, or general world-generation objectives, rather than on data that explicitly couples mid-rollout event injection with off-screen subject evolution. Their behavior in our setting should therefore not be interpreted as a general assessment of their visual quality or intended capabilities. We leave systematic adaptation and controlled comparison of these models to future work.

\paragraph{Evaluated baselines.}
For LingBot-World v1 \citep{lingbot-world-v1} and v2 \citep{lingbot-world-v2}, we use the official interactive interfaces, translate the benchmark trajectory into their native camera controls, and schedule all prompts within one persistent rollout. We evaluate the released YUME-1.5 5B checkpoint \citep{mao2026yume1.5} at 720p through its long-context autoregressive interface. For SANA-WM \citep{zhu2026sana-wm}, we use the released streaming checkpoint and assign the three prompts to a $3/2/3$ schedule over its eight autoregressive blocks, with camera motion supplied through its native trajectory interface. Echo-WM \citep{zhang2026echowm} is evaluated with sequential prompt switching while retaining its persistent autoregressive world state. Evoke and Evoke-Turbo \citep{yin2026alaya-evoke} use the official image-to-video segment mode with matched inputs and prompt boundaries, differing only in their released variants and native sampling configurations.

\paragraph{One-shot references.}\label{app:oneshot}
We additionally evaluate Wan2.2 TI2V-5B \citep{wan2025wan}, LTX-2.3, and LTX-2.5 \citep{hacohen2026ltx-2} as general one-shot video generators. Each model receives a single composed prompt containing the initial, event, and blind revisit instructions and generates the complete video in one pass. These models therefore observe the intended event and camera-return instruction before generation begins, unlike the main autoregressive setting in which the event is introduced only during rollout. We report their results in \cref{tab:oneshot-references} for completeness but do not include them in the main comparison.

\begin{table}[h]
\centering
\caption{One-shot general video generators on \ours-Bench, with \ours\ included as an autoregressive reference below the separator. Best results among the one-shot models are bold (\%).}
\label{tab:oneshot-references}
\scriptsize
\setlength{\tabcolsep}{3.0pt}
\renewcommand{\arraystretch}{1.12}
\resizebox{\textwidth}{!}{%
\begin{tabular}{lccccccccc}
\toprule
& \multicolumn{5}{c}{Task metrics} & \multicolumn{4}{c}{Visual quality, consistency, and motion} \\
\cmidrule(lr){2-6}\cmidrule(lr){7-10}
Method & Core & Revisit & Event & Causal & React. & Visual & Subj. Cons. & Bg. Cons. & Motion \\
\midrule
Wan2.2 TI2V-5B \citep{wan2025wan} & 8.3 & 40.0 & 87.5 & 17.1 & 18.3 & 77.9 & 81.7 & 95.4 & 98.8 \\
LTX-2.3 \citep{hacohen2026ltx-2} & 27.5 & \textbf{97.5} & \textbf{94.2} & 43.3 & 42.1 & 88.8 & \textbf{96.7} & \textbf{99.2} & 99.2 \\
LTX-2.5 \citep{hacohen2026ltx-2} & \textbf{34.2} & 95.0 & 90.8 & \textbf{50.8} & \textbf{42.9} & \textbf{90.4} & 95.8 & 98.8 & \textbf{99.4} \\
\midrule
\rowcolor{oursrow}
\textcolor{gray}{\textbf{\ours\ (ours; autoregressive)}} & \textcolor{gray}{90.8} & \textcolor{gray}{100.0} & \textcolor{gray}{97.5} & \textcolor{gray}{94.2} & \textcolor{gray}{85.0} & \textcolor{gray}{83.8} & \textcolor{gray}{99.2} & \textcolor{gray}{98.3} & \textcolor{gray}{99.5} \\
\bottomrule
\end{tabular}%
}
\end{table}

\begin{wraptable}{r}{0.56\textwidth}
\centering
\caption{Task metrics of one-shot MiniMax-H3 on \ours-Bench.}
\label{tab:h3-oneshot}
\scriptsize
\setlength{\tabcolsep}{3.0pt}
\renewcommand{\arraystretch}{1.12}
\resizebox{\linewidth}{!}{%
\begin{tabular}{lccccc}
\toprule
Method & Core & Revisit & Event & Causal & React. \\
\midrule
MiniMax-H3 \citep{minimax2026h3} & 21.7 & 100.0 & 99.2 & 44.2 & 47.9 \\
\midrule
\rowcolor{oursrow}
\textcolor{gray}{\textbf{\ours\ (ours; autoregressive)}} & \textcolor{gray}{90.8} & \textcolor{gray}{100.0} & \textcolor{gray}{97.5} & \textcolor{gray}{94.2} & \textcolor{gray}{85.0} \\
\bottomrule
\end{tabular}%
}
\vspace{-0.5\baselineskip}
\end{wraptable}

We further evaluate MiniMax-H3~\citep{minimax2026h3} ($\sim$33B) on EvoWorld-Bench based on the one-shot setting. As shown in \cref{tab:h3-oneshot}, its high Event and Revisit scores do not translate into comparable Core task performance; successful components alone do not guarantee event-conditioned causal evolution.
\par
\vspace{0.25\baselineskip}
\WFclear

\section{Limitations and Future Work}\label{app:limitations}

Although the segment-autoregressive design does not impose a fixed semantic horizon and can autoregressively extend the rollout, the present work focuses on a single event followed by a single subject reaction. This controlled scope isolates the minimal out-of-sight event--reaction chain, but does not yet capture long-horizon worlds in which multiple events interact, earlier reactions alter later conditions, or several entities respond over time. Extending the data, reasoning memory, training procedure, and evaluation protocol to multi-event and multi-reaction evolution is therefore an important direction for future work. In addition, our current scenarios primarily employ horizontal move-away and move-back camera trajectories because they represent the most common and directly interpretable camera motion for revisit scenarios. We do not systematically evaluate more complex camera motion, such as large vertical displacement, depth-wise movement, orbiting, or compound six-degree-of-freedom trajectories, which may make scene correspondence, entity retrieval, and revisit consistency substantially more difficult. Future versions of \ours-Data and \ours-Bench should incorporate these trajectories and evaluate whether the framework remains robust under more general interactive camera control. 


\paragraph{Acknowledgements.}
We thank all collaborators for their support and contributions throughout this project. We are also grateful to Su Guo, Jingyang Jia, and Shengrong Gu for their helpful suggestions on this work.

\end{document}